\documentclass[10pt,twocolumn,letterpaper]{article}

\usepackage{cvpr}
\usepackage{times}
\usepackage{epsfig}
\usepackage{graphicx}
\usepackage{amsmath}
\usepackage{amssymb}
\usepackage{enumerate}
\usepackage{cite}
\usepackage{caption}
\usepackage{subcaption}

\DeclareMathOperator*{\argmax}{argmax}

\newcommand{\ignore}[1]{}

\usepackage{color}

\newcommand{\QP}[1]{{${\bf QP}_{#1}$}}

\usepackage[breaklinks=true,bookmarks=false]{hyperref}

\cvprfinalcopy 

\def\cvprPaperID{261} 

\begin{document}

\title{Bottom-Up and Top-Down Reasoning with Hierarchical Rectified Gaussians}

\author{Peiyun Hu\\
UC Irvine\\
{\tt\small peiyunh@ics.uci.edu}
\and
Deva Ramanan\\
Carnegie Mellon University\\
{\tt\small dramanan@cs.cmu.edu}
}

\maketitle

\begin{abstract}
  Convolutional neural nets (CNNs) have demonstrated remarkable
  performance in recent history. Such approaches tend to work in a
  ``unidirectional'' bottom-up feed-forward fashion. However,
  practical experience and biological evidence tells us that feedback plays a crucial role,
  particularly for detailed spatial understanding tasks. This work
  explores ``bidirectional'' architectures that also reason with
  top-down feedback: neural units are influenced by both lower and
  higher-level units.

  We do so by treating units as rectified latent variables in a
  quadratic energy function, which can be seen as a hierarchical
  Rectified Gaussian model (RGs)~\cite{socci1998rectified}. We show
  that RGs can be optimized with a quadratic program (QP), that can in
  turn be optimized with a recurrent neural network (with rectified
  linear units). This allows RGs to be trained with GPU-optimized
  gradient descent. From a theoretical perspective, RGs help establish
  a connection between CNNs and hierarchical probabilistic
  models. From a practical perspective, RGs are well suited for
  detailed spatial tasks that can benefit from top-down reasoning. We
  illustrate them on the challenging task of keypoint localization
  under occlusions, where local bottom-up evidence may be
  misleading. We demonstrate state-of-the-art results on challenging
  benchmarks.

\end{abstract}

\section{Introduction}
\begin{figure}[t!]
  \centering
  \includegraphics[width=\linewidth]{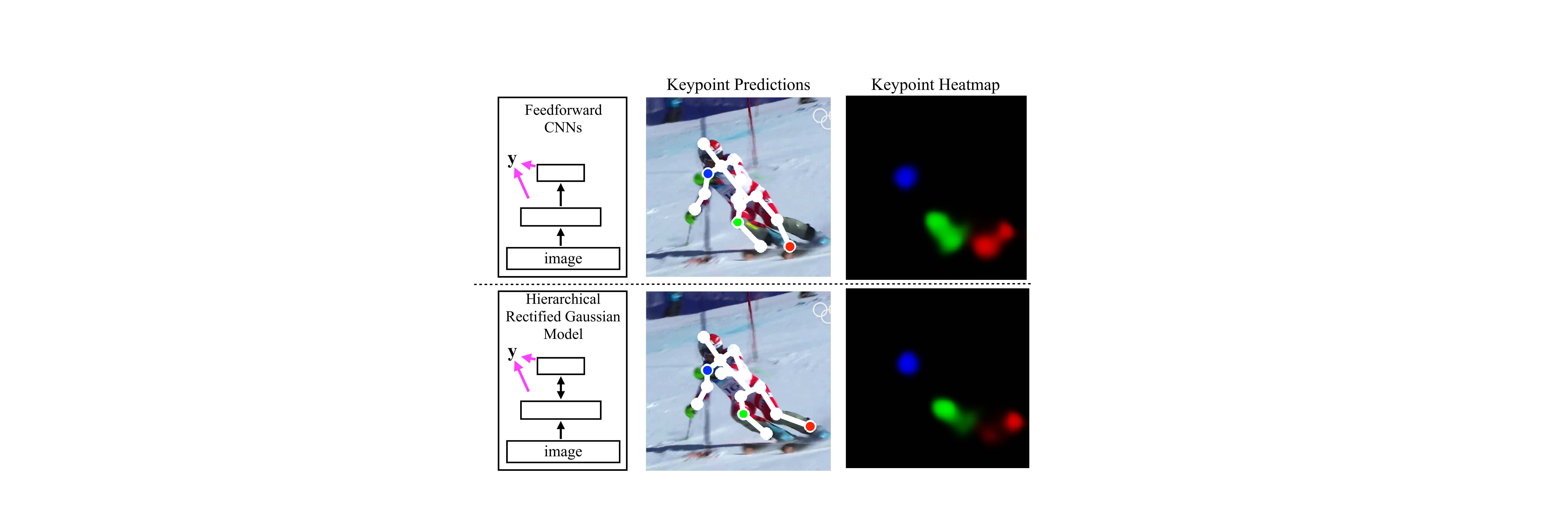}
  \caption{On the {\bf top}, we show a state-of-the-art multi-scale
    feedforward net, trained for keypoint heatmap prediction, where
    the blue keypoint (the right shoulder) is visualized in the blue
    plane of the RGB heatmap. The ankle keypoint (red) is confused
    between left and right legs, and the knee (green) is poorly
    localized along the leg. We believe this confusion arises from
    bottom-up computations of neural activations in a feedforward
    network. On the {\bf bottom}, we introduce hierarchical Rectified
    Gaussian (RG) models that incorporate top-down feedback by
    treating neural units as latent variables in a quadratic energy
    function. Inference on RGs can be unrolled into recurrent nets
    with rectified activations. Such architectures produce better
    features for ``vision-with-scrutiny''
    tasks~\cite{hochstein2002view} (such as keypoint prediction)
    because lower-layers receive top-down feedback from above. Leg
    keypoints are much better localized with top-down knowledge (that
    may capture global constraints such as kinematic
    consistency).}
  \label{fig:splash}
\end{figure}
Hierarchical models of visual processing date back to the iconic work
of Marr~\cite{marr1982vision}.  Convolutional neural nets (CNN's),
pioneered by LeCun {\em et al.}~\cite{lecun1998gradient}, are
hierarchical models that compute progressively more invariant
representations of an image in a bottom-up, feedforward fashion. They
have demonstrated remarkable progress in recent history for visual
tasks such as classification~\cite{krizhevsky2012imagenet,
  simonyan2014very, szegedy2014going}, object
detection~\cite{girshick2014rich}, and image
captioning~\cite{karpathy2014deep}, among others.

{\bf Feedback in biology:} 
Biological evidence suggests that {\em vision at a glance} tasks, such
as rapid scene categorization~\cite{vanrullen2001bird}, can be
effectively computed with feedforward hierarchical
processing. However, {\em vision with scrutiny} tasks, such as
fine-grained categorization~\cite{kosslyn1995topographical} or
detailed spatial manipulations~\cite{ito1999attention}, appear to
require feedback along a ``reverse
hierarchy''~\cite{hochstein2002view}. Indeed, most neural connections
in the visual cortex are believed to be feedback rather than
feedforward~\cite{douglas1995recurrent, kruger2013deep}. 

{\bf Feedback in computer vision:} Feedback has also played a central
role in many classic computer vision models. Hierarchical
probabilistic models~\cite{zhu2011recursive, jin2006context,
  lee2003hierarchical}, allow random variables in one layer to be
naturally influenced by those above and below. For example, lower
layer variables may encode edges, middle layer variables may encode
parts, while higher layers encode objects. Part
models~\cite{felzenszwalb2010object} allow a face object to influence
the activation of an eye part through top-down feedback, which is
particularly vital for occluded parts that receive misleading
bottom-up signals. Interestingly, feed-forward
inference on part models can be written as a
CNN~\cite{girshick2014deformable}, but the proposed mapping does not
hold for feedback inference. 

{\bf Overview:} To endow CNNs with feedback, we treat neural units as
nonnegative latent variables in a quadratic energy function. When
probabilistically normalized, our quadratic energy function
corresponds to a Rectified Gaussian (RG) distribution, for which
inference can be cast as a quadratic program
(QP)~\cite{socci1998rectified}. We demonstrate that coordinate descent
optimization steps of the QP can be ``unrolled'' into a recurrent
neural net with rectified linear units. This observation allows us to
discriminatively-tune RGs with neural network toolboxes: {\em we tune
  Gaussian parameters such that, when latent variables are inferred
  from an image, the variables act as good features for discriminative
  tasks}. From a theoretical perspective, RGs help establish a
connection between CNNs and hierarchical probabilistic models. From a
practical perspective, we introduce RG variants of state-of-the-art
deep models (such as VGG16~\cite{simonyan2014very}) that require no
additional parameters, but consistently improve performance due to the
integration of top-down knowledge.

\section{Hierarchical Rectified Gaussians} \label{sec:lvm} In this
section, we describe the Rectified Gaussian models of Socci and
Seung~\cite{socci1998rectified} and their relationship with rectified
neural nets. Because we will focus on convolutional nets, it will help
to think of variables $z = [z_i]$ as organized into layers, spatial
locations, and channels (much like the neural activations of a CNN).
We begin by defining a quadratic energy over variables
$z$:
\begin{align}
  S(z) &= \frac{1}{2} z^T W z + b^Tz \label{eq:score}\\
  P(z) &\propto e^{S(z)} \nonumber\\
  \text{Boltzmann:} & \quad z_i \in \{0,1\}, w_{ii} = 0 \nonumber\\
  \text{Gaussian:} & \quad z_i \in R, -W \text{ is PSD} \nonumber\\
  \text{Rect. Gaussian:} & \quad z_i \in R^+, -W \text{ is copositive} \nonumber
\end{align}
where $W = [w_{ij}]$, $b=[b_i]$. The symmetric matrix $W$ captures
bidirectional interactions between low-level features (e.g., edges)
and high-level features (e.g., objects). Probabilistic models such as
Boltzmann machines, Gaussians, and Rectified Gaussians differ simply
in restrictions on the latent variable - binary, continuous, or
nonnegative. Hierarchical models, such as deep Boltzmann
machines~\cite{salakhutdinov2009deep}, can be written as a special
case of a block-sparse matrix $W$ that ensures that only neighboring
layers have direct interactions.

\begin{figure}[t!]
  \centering
  \includegraphics[width=0.75\linewidth]{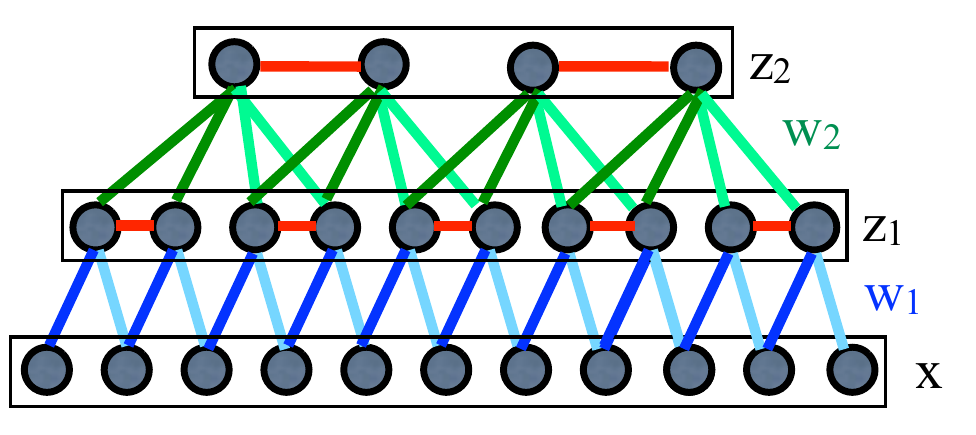}
  \caption{A hierarchical Rectified Gaussian model where latent
    variables $z_i$ are denoted by circles, and arranged into layers and
    spatial locations. We write $x$ for the input image and $w_i$ for
    convolutional weights connecting layer $i-1$ to $i$. Lateral
    inhibitory connections between latent variables are drawn in
    red. Layer-wise coordinate updates are computed by filtering,
    rectification, and non-maximal suppression.}
  \label{fig:rlvm}
\end{figure}

{\bf Normalization:} To ensure that the scoring function can be
probabilistically normalized, Gaussian models require that $(-W)$ be
positive semidefinite (PSD) ($-z^TWz \geq 0, \forall
z$) 
Socci and Seung~\cite{socci1998rectified} show that Rectified
Gaussians require the matrix $(-W)$ to only be {\em copositive}
(-$z^TWz \geq 0, \forall z \geq 0$), which is a strictly weaker
condition. Intuitively, copositivity ensures that the maximum of
$S(z)$ is still finite, allowing one to compute the partition
function. This relaxation significantly increases the expressive power
of a Rectified Gaussian, allowing for multimodal distributions. We
refer the reader to the excellent discussion
in~\cite{socci1998rectified} for further details.

{\bf Comparison:} Given observations (the image) in the lowest layer,
we will infer the latent states (the features) from the above
layers. Gaussian models are limited in that features will always be
linear functions of the image. Boltzmann machines produce nonlinear
features, but may be limited in that they pass only binary information
across layers~\cite{nair2010rectified}. Rectified Gaussians are
nonlinear, but pass continuous information across layers: $z_i$
encodes the presence or absence of a feature, and if present, the
strength of this activation (possibly emulating the firing rate of a
neuron~\cite{kandel2000principles}).

{\bf Inference:} Socci and Seung point out that MAP estimation of
Rectified Gaussians can be formulated as a quadratic program (QP) with
nonnegativity constraints~\cite{socci1998rectified}:
\begin{align}
  \max_{z \geq 0} \frac{1}{2}z^TWz + b^Tz \label{eq:qp}
\end{align}
However, rather than using projected gradient descent (as proposed
by~\cite{socci1998rectified}), we show that coordinate descent is
particularly effective in exploiting the sparsity of
$W$. Specifically, let us optimize a single $z_i$ holding all others
fixed. Maximizing a 1-d quadratic function subject to non-negative
constraints is easily done by solving for the optimum and clipping:
\begin{align}
  \max_{z_i \geq 0} f(z_i) &\quad \text{where} \quad f(z_i) = \frac{1}{2} w_{ii} z_i^2 + (b_i + \sum_{j \neq i} w_{ij}z_j)z_i \nonumber\\
  \frac{\partial f}{\partial z_i}  &=  w_{ii}z_i + b_i + \sum_{j \neq i} w_{ij}z_j = 0 \nonumber\\
  z_i &= -\frac{1}{w_{ii}}\max(0,b_i + \sum_{j \neq i} w_{ij} z_j) \label{eq:coor}\\
  &= \max(0,b_i + \sum_{j \neq i} w_{ij} z_j)  \quad \text{for} \quad  w_{ii} = -1 \nonumber
\end{align}
By fixing $w_{ii} = -1$ (which we do for all our experiments), the above maximization can solved with a rectified dot-product operation.

{\bf Layerwise-updates:} The above updates can be performed for all latent variables in a layer in parallel. With a slight abuse of
notation, let us define the input image to be the (observed) bottom-most
layer $x=z_0$, and the variable at layer $i$ and spatial position $u$
is written as $z_i[u]$. The weight connecting $z_{i-1}[v]$ to
$z_{i}[u]$ is given by $w_i[\tau]$, where $\tau = u - v$ depends only
on the relative offset between $u$ and $v$ (visualized in
Fig.~\ref{fig:rlvm}):
\begin{align}
  z_i[u] &= \max(0, b_i + top_i[u] + bot_i[u]) \label{eq:layer} \quad \text{where}\\
  top_i[u] &= \sum_{\tau} w_{i+1}[\tau] z_{i+1} [u-\tau] \nonumber\\
  bot_i[u] &= \sum_{\tau} w_{i}[\tau] z_{i-1}[u+\tau] \nonumber
\end{align}
\noindent where we assume that layers have a single one-dimensional
channel of a fixed length to simplify notation. By tying together
weights such that they only depend on relative locations, bottom-up
signals can be computed with cross-correlational filtering, while
top-down signals can be computed with convolution. In the existing
literature, these are sometimes referred to as deconvolutional and
convolutional filters (related through a $180^\circ$
rotation)~\cite{zeiler2010deconvolutional}. It is natural to start
coordinate updates from the bottom layer $z_1$, initializing all
variables to 0. During the initial bottom-up coordinate pass, $top_i$
will always be 0. This means that the bottom-up coordinate updates can
be computed with simple filtering and thresholding. {\em Hence a
  single bottom-up pass of layer-wise coordinate optimization of a
  Rectified Gaussian model can be implemented with a CNN.}

\begin{figure}[t!]
  \centering
  \includegraphics[width=0.98\linewidth]{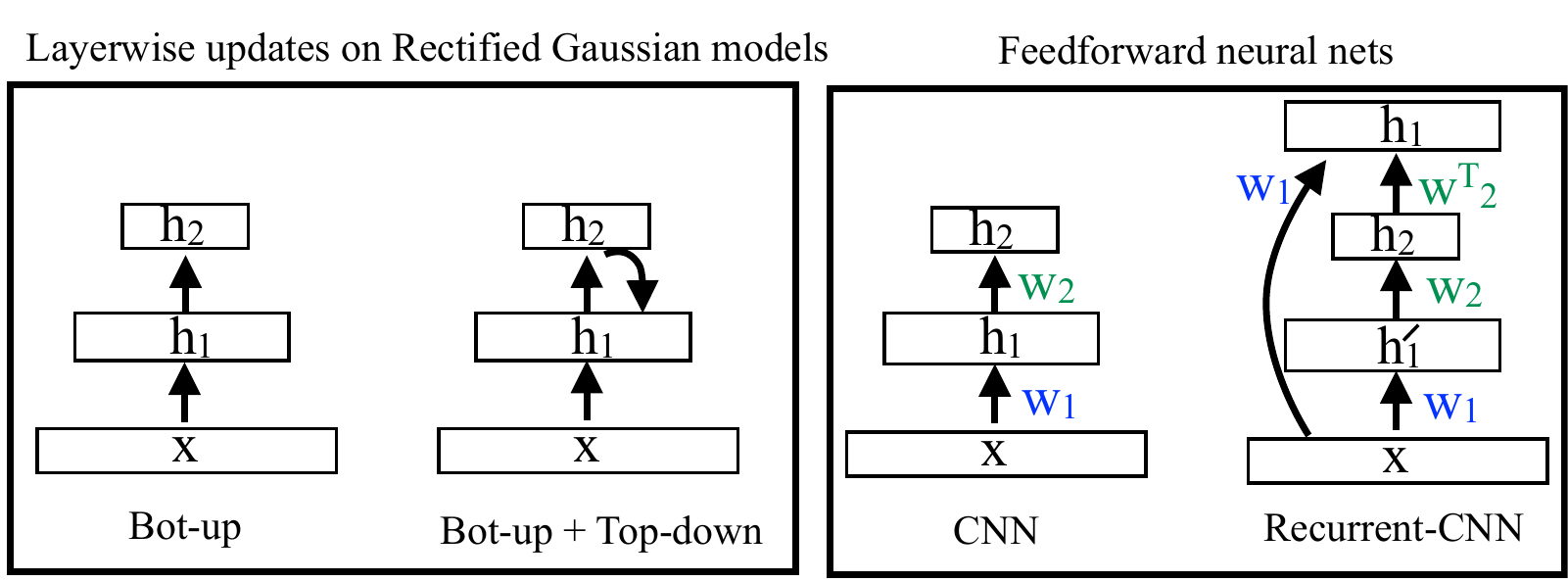}
  \caption{On the {\bf left}, we visualize two sequences of layer-wise
    coordinate updates on our latent-variable model. The first is a
    bottom-up pass, while the second is a bottom-up + top-down
    pass. On the {\bf right}, we show that bottom-up updates can be
    computed with a feed-forward CNN, and bottom-up-and-top-down
    updates can be computed with an ``unrolled'' CNN with additional
    skip connections and tied weights (which we define as a recurrent
    CNN). We use $^T$ to denote a $180^\circ$ rotation of filters that
    maps correlation to convolution. We follow the color scheme from
    Fig.~\ref{fig:rlvm}.
  }
  \label{fig:info}
\end{figure}

{\bf Top-down feedback:} We add top-down feedback simply by applying
additional coordinate updates \eqref{eq:layer} in a top-down fashion,
from the top-most layer to the bottom. Fig.~\ref{fig:info} shows that
such a sequence of bottom-up and top-down updates can be ``unrolled''
into a feed-forward CNN with ``skip'' connections between layers and
tied weights. One can interpret such a model as a recurrent CNN that
is capable of feedback, since lower-layer variables (capturing say,
edges) can now be influenced by the activations of high-layer
variables (capturing say, objects).  Note that we make use of
recurrence along the depth of the hierarchy, rather than along time or
spacial dimensions as is typically done~\cite{haykin2009neural}. When
the associated weight matrix $W$ is copositive, an infinitely-deep
recurrent CNN {\em must} converge to the solution of the QP from
\eqref{eq:qp}.

{\bf Non-maximal suppression (NMS):} To encourage sparse activations,
we add lateral inhibitory connections between variables from same
groups in a layer. Specifically, we write the weight connecting
$z_i[u]$ and $z_i[v]$ for $(u,v) \in \text{group}$ as
$w_i[u,v] = -\infty$. Such connections are shown as red edges in
Fig.~\ref{fig:rlvm}. For disjoint groups (say, non-overlapping 2x2
windows), {\em layer-wise updates correspond to filtering,
  rectification \eqref{eq:layer}, and non-maximal suppression (NMS)
  within each group.}

Unlike max-pooling, NMS encodes the spatial location of the max by
returning 0 values for non-maximal locations. Standard max-pooling can
be obtained as a special case by replicating filter weights $w_{i+1}$
across variables $z_i$ within the same group (as shown in
Fig.~\ref{fig:rlvm}). This makes NMS independent of the top-down
signal $top_i$. However, our approach is more general in that NMS can
be guided by top-down feedback: high-level variables (e.g., car detections)
influence the spatial location of low-level
variables (e.g., wheels), which is particularly helpful when parsing occluded wheels. Interestingly, top-down feedback seems to encode
spatial information without requiring additional ``capsule''
variables~\cite{hinton2011transforming}.

\begin{figure*}[t]
  \centering
  \includegraphics[width=\linewidth]{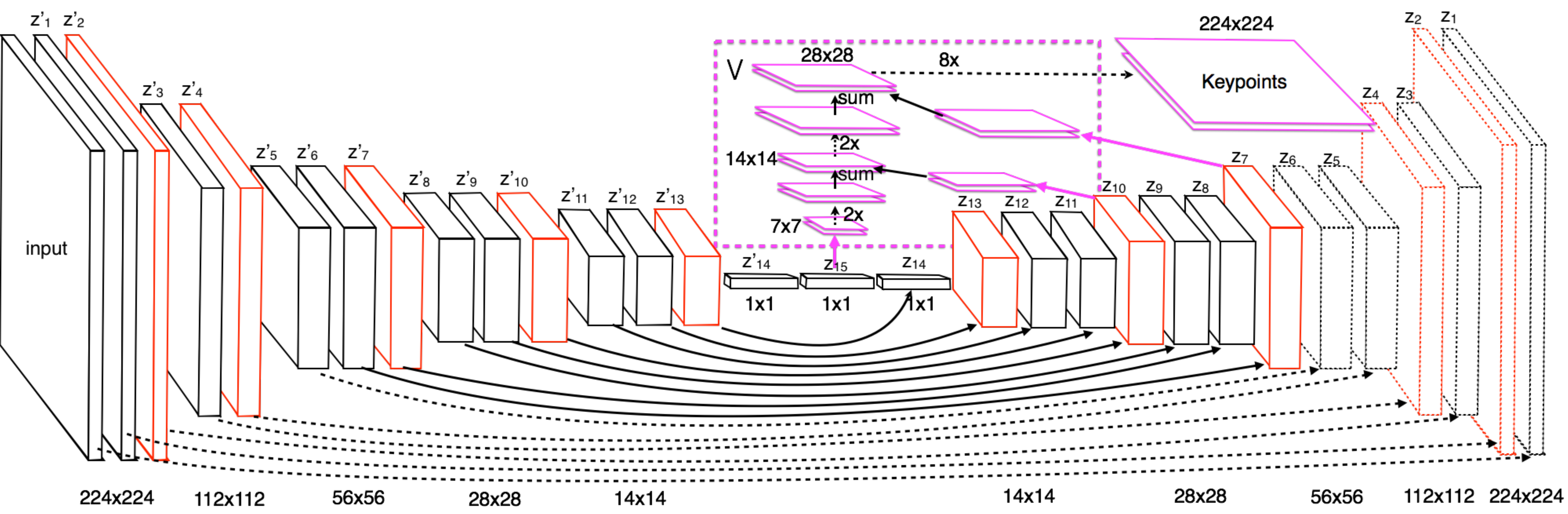}\\
  \caption{We show the architecture of \QP{2} implemented in our
    experiments. \QP{1} corresponds to the left half of \QP{2}, which
    essentially resembles the state-of-the-art VGG-16 CNN
    \cite{simonyan2014very}. \QP{2} is implemented with an 2X
    ``unrolled'' recurrent CNN with transposed weights, skip
    connections, and zero-interlaced upsampling (as shown in
    Fig.~\ref{fig:cnn}). Importantly, \QP{2} does not require any
    additional parameters. Red layers include lateral inhibitory
    connections enforced with NMS. Purple layers denote multi-scale
    convolutional filters that (linearly) predict keypoint heatmaps
    given activations from different layers. Multi-scale filters are
    efficiently implemented with coarse-to-fine
    upsampling~\cite{long2014fully}, visualized in the purple dotted
    rectangle (to reduce clutter, we visualize only 3 of the 4 multiscale layers). Dotted layers are not implemented to reduce memory.}
  \label{fig:arch}
\end{figure*}

{\bf Approximate inference:} Given the above global scoring function
and an image $x$, inference corresponds to $\argmax_zS(x,z)$. As
argued above, this can be implemented with an infinitely-deep unrolled
recurrent CNN. However, rather than optimizing the latent variables to
completion, we perform a fixed number ($k$) of layer-wise coordinate
descent updates. This is guaranteed to report back finite variables
$z^*$ for any weight matrix $W$ (even when not copositive):
\begin{align}
  z^* = {\bf QP}_k(x,W,b) \label{eq:infer}, \quad z^* \in R^N
\end{align}
We write ${\bf QP}_k$ in bold to emphasize that it is a {\em
  vector-valued function} implementing $k$ passes of layer-wise
coordinate descent on the QP from \eqref{eq:qp}, returning a vector of
all $N$ latent variables. We set $k=1$ for a single bottom-up pass
(corresponding to a standard feed-forward CNN) and $k=2$ for an
additional top-down pass. We visualize examples of recurrent CNNs that
implement \QP{1} and \QP{2} in Fig.~\ref{fig:arch}.

{\bf Output prediction:} We will use these $N$ variables as features
for $M$ recognition tasks. In our experiments, we consider the task of
predicting heatmaps for $M$ keypoints. Because our latent variables
serve as rich, multi-scale description of image features, we assume
that simple linear predictors built on them will suffice:
\begin{align}
  y &= V^T z^*, \quad y \in R^M, V \in R^{N \times M}
\end{align}

{\bf Training:} Our overall model is parameterized by
$(W,V,b)$. Assume we are given training data pairs of images and
output label vectors $\{x_i,y_i\}$. We define a training objective as
follows
\begin{align}
  \min_{W,V,b} R(W) + R(V) + \sum_i \text{loss}(y_i,V^T {\bf QP}_k(x_i,W,b)) \label{eq:obj}
\end{align}
\noindent where $R$ are regularizer functions (we use the Frobenius
matrix norm) and ``loss" sums the loss of our $M$ prediction tasks
(where each is scored with log or softmax loss). We optimize the above
by stochastic gradient descent. Because ${\bf QP}_k$ is a
deterministic function, its
gradient 
with respect to $(W,b)$ can be computed by backprop on the $k$-times
unrolled recurrent CNN (Fig.~\ref{fig:info}). We choose to separate
$V$ from $W$ to ensure that feature extraction does not scale with the
number of output tasks (${\bf QP}_k$ is independent of $M$).  During
learning, we fix diagonal weights $(w_{i}[u,u] = -1)$ and lateral
inhibition weights ($w_i[u,v] = -\infty$ for
$(u,v) \in \text{group}$).

{\bf Related work (learning):} The use of gradient-based
backpropagation to learn an unrolled model dates back to
`backprop-through-structure'
algorithms~\cite{goller1996learning,socher2011parsing} and graph
transducer networks~\cite{lecun1998gradient}. More recently, such
approaches were explored general graphical
models~\cite{stoyanov2011empirical} and Boltzmann
machines~\cite{goodfellow2013multi}. Our work uses such ideas to learn
CNNs with top-down feedback using an unrolled latent-variable
model.

{\bf Related work (top-down):} Prior work has explored networks that
reconstruct images given top-down cues. This is often cast as
unsupervised learning with
autoencoders~\cite{hinton2006reducing,vincent2010stacked,masci2011stacked}
or deconvolutional networks~\cite{zeiler2010deconvolutional}, though
supervised variants also exist
~\cite{long2014fully,noh2015learning}. Our network differs in that all
nonlinear operations (rectification and max-pooling) are influenced by
both bottom-up and top-down knowledge~\eqref{eq:layer}, which is
justified from a latent-variable perspective.

\section{Implementation}
\label{sec:imp}
In this section, we provide details for implementing \QP{1} and
\QP{2} with existing CNN toolboxes. We visualize our specific
architecture in Fig.~\ref{fig:arch}, which closely follows the
state-of-the-art VGG-16 network~\cite{simonyan2014very}. We use 3x3
filters and 2x2 non-overlapping pooling windows (for NMS). Note that,
when processing NMS-layers, we conceptually use 6x6 filters with
replication after NMS, which in practice can be implemented with
standard max-pooling and 3x3 filters (as argued in the previous
section). Hence \QP{1} {\em is} essentially a re-implementation of
VGG-16.

\begin{figure}[t!]
  \centering
  \includegraphics[width=1\linewidth]{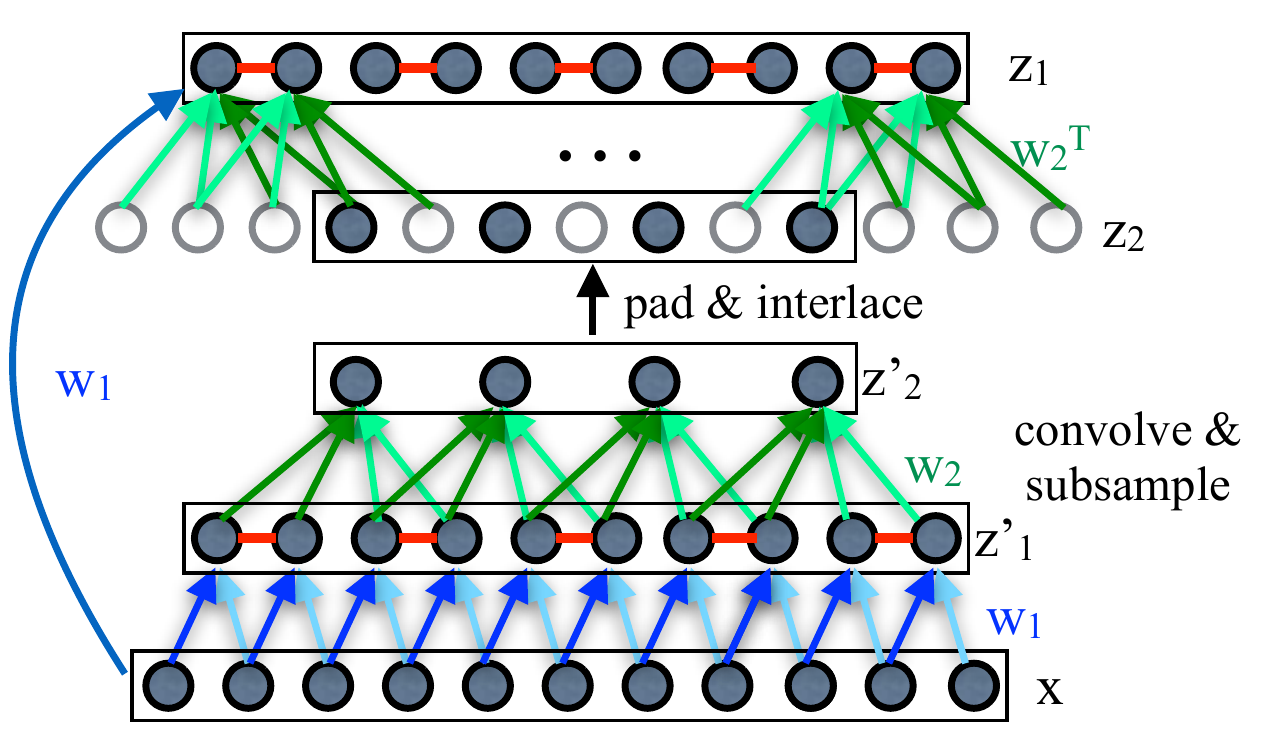}
  \caption{Two-pass layer-wise coordinate descent for a two-layer Rectified Gaussian
    model can be implemented with modified CNN operations. White
    circles denote 0's used for interlacing and border padding. We
    omit rectification operations to reduce clutter. We follow the
    color scheme from Fig.~\ref{fig:rlvm}. }
  \label{fig:cnn}
\end{figure}

{\bf \QP{2}:} Fig.~\ref{fig:cnn} illustrates top-down coordinate
updates, which require additional feedforward layers, skip
connections, and tied weights. Even though \QP{2} is twice as deep
as \QP{1} (and~\cite{simonyan2014very}), {\em it requires no
  additional parameters}. Hence top-down reasoning ``comes for
free''. There is a small notational inconvenience at layers that
decrease in size. In typical CNNs, this decrease arises from a
previous pooling operation. Our model requires an explicit $2\times$
subsampling step (sometimes known as strided filtering) because it
employs NMS instead of max-pooling. When this subsampled layer is
later used to produce a top-down signal for a future coordinate
update, variables must be zero-interlaced before applying the
$180^\circ$ rotated convolutional filters (as shown by hollow circles
in Fig.~\ref{fig:cnn}). Note that is {\em not} an approximation, but
the mathematically-correct application of coordinate descent given
subsampled weight connections.

{\bf Supervision $y$:} The target label for a single keypoint is a
sparse 2D heat map with a `1' at the keypoint location (or all `0's if
that keypoint is not visible on a particular training image). We score
this heatmap with a per-pixel log-loss. In practice, we assign `1's to
a circular neighborhood that implicitly adds jittered keypoints to the
set of positive examples. 

{\bf Multi-scale classifiers $V$:} We implement our output
classifiers~\eqref{eq:obj} as multi-scale convolutional filters
defined over different layers of our model. We use upsampling to
enable efficient coarse-to-fine computations, as described for
fully-convolutional networks (FCNs)~\cite{long2014fully} (and shown in
Fig.~\ref{fig:arch}). Specifically, our multi-scale filters are
implemented as $1\times1$ filters over 4 layers (referred to as fc7, pool4,
pool3, and pool2 in \cite{simonyan2014very}). Because our top (fc7)
layer is limited in spatial resolution (1x1x4096), we define our
coarse-scale filter to be ``spatially-varying'', which can
alternatively be thought of as a linear ``fully-connected'' layer that
is reshaped to predict a coarse (7x7) heatmap of keypoint predictions
given fc7 features. Our intuition is that spatially-coarse global
features can still encode global constraints (such as viewpoints) that
can produce coarse keypoint predictions. This coarse predictions are
upsampled and added to the prediction from pool4, and so on (as in
~\cite{long2014fully}).

{\bf Multi-scale training:} We initialize parameters of both \QP{1}
and \QP{2} to the pre-trained VGG-16 model\cite{simonyan2014very}, and
follow the coarse-to-fine training scheme for learning
FCNs~\cite{long2014fully}. Specifically, we first train coarse-scale
filters, defined on high-level (fc7) variables.
Note that \QP{1} and \QP{2} are equivalent in this setting. This
coarse-scale model is later used to initialize a two-scale predictor,
where now \QP{1} and \QP{2} differ. 
The process is repeated up until the full multi-scale model is
learned. To save memory during various stages of learning, we only
instantiate \QP{2} up to the last layer used by the multi-scale
predictor (not suitable for \QP{k} when $k>2$). We use a batch size of
40 images, a fixed learning rate of $10^{-6}$, momentum of 0.9 and
weight decay of 0.0005. We also decrease learning rates of parameters
built on lower scales \cite{long2014fully} by a factor of 10. Batch
normalization\cite{ioffe2015batch} is used before each
non-linearity. Both our models and code are
available online \footnote{\url{https://github.com/peiyunh/rg-mpii}}.

{\bf Prior work:} We briefly compare our approach to recent work on
keypoint prediction that make use of deep architectures.  Many
approaches incorporate multi-scale cues by evaluating a deep network
over an image pyramid~\cite{tulsiani2014viewpoints,
  tompson2014efficient, tompson2014joint}. Our model processes only a
single image scale, extracting multi-scale features from multiple
layers of a single network, where importantly, fine-scale features are
refined through top-down feedback. Other approaches cast the problem
as one of regression, where (x,y) keypoint locations are
predicted~\cite{zhang2014facial} and often iteratively
refined~\cite{carreira2015human,sun2013deep}. Our models predict
heatmaps, which can be thought of as {\em marginal distributions} over
the (x,y) location of a keypoint, capturing uncertainty. We show that
by thresholding the heatmap value (certainty), one can also produce
{\em keypoint visibility} estimates ``for free''. Our comments hold
for our bottom-up model \QP{1}, which can be thought of as a FCN tuned
for keypoint heatmap prediction, rather than semantic pixel
labeling. Indeed, we find such an approach to be a surprisingly simple
but effective baseline that outperforms much prior work.

\section{Experiment Results}

We evaluated fine-scale keypoint localization on several benchmark
datasets of human faces and bodies. To better illustrate the benefit
of top-down feedback, we focus on datasets with significant
occlusions, where bottom-up cues will be less reliable. All datasets
provide a rough detection window for the face/body of interest. We
crop and resize detection windows to 224x224 before feeding into our
model. Recall that \QP{1} is essentially a re-implementation of a
FCN~\cite{long2014fully} defined on a VGG-16
network~\cite{simonyan2014very}, and so represents quite a strong
baseline.  Also recall that \QP{2} adds top-down reasoning {\em
  without any increase in the number of parameters}. We will show this
consistently improves performance, sometimes considerably.  Unless
otherwise stated, results are presented for a 4-scale multi-scale
model.

\begin{figure}[t!]
  \centering
  \begin{subfigure}{.75\linewidth}
    \includegraphics[width=\linewidth]{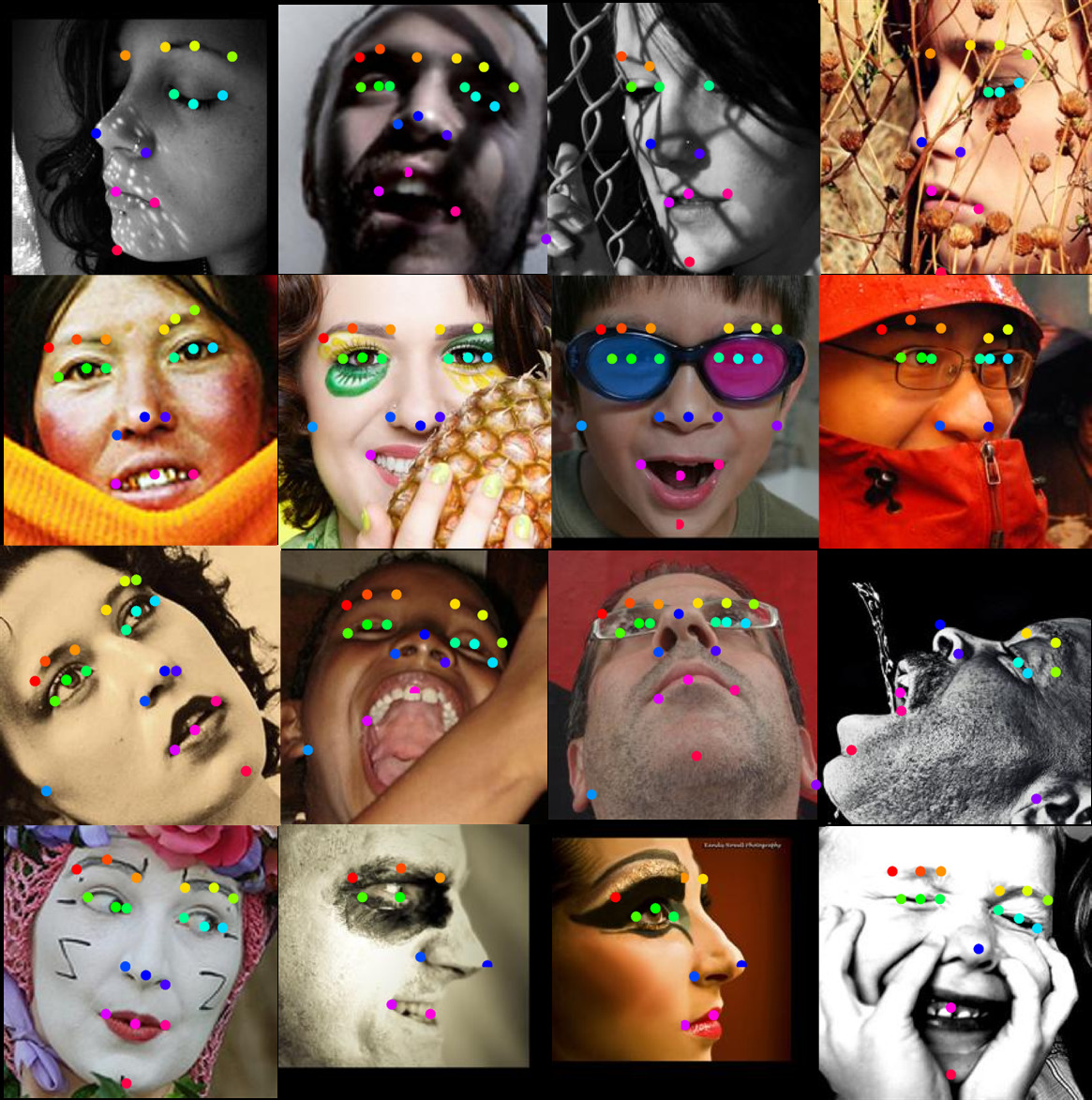}
    \caption{}
  \end{subfigure}
  \begin{subfigure}{.195\linewidth}
    \includegraphics[width=\linewidth]{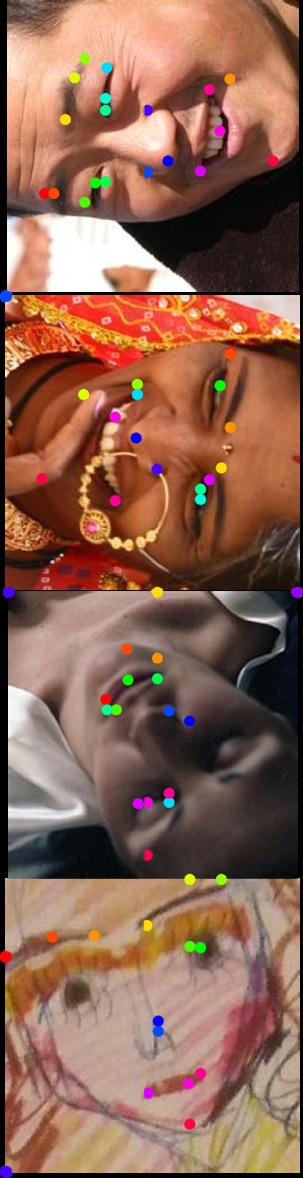}
    \caption{}
  \end{subfigure}
  \caption{Facial landmark localization results of \QP{2} on AFLW,
    where landmark ids are denoted by color. We only plot landmarks
    annotated visible. Our bidirectional model is able to deal with
    large variations in illumination, appearance and pose ({\bf
      a}). We show images with multiple challenges present in ({\bf
      b}).}
  \label{fig:aflw}
\end{figure}

\begin{figure}[t!]
  \centering
\includegraphics[width=.75\linewidth]{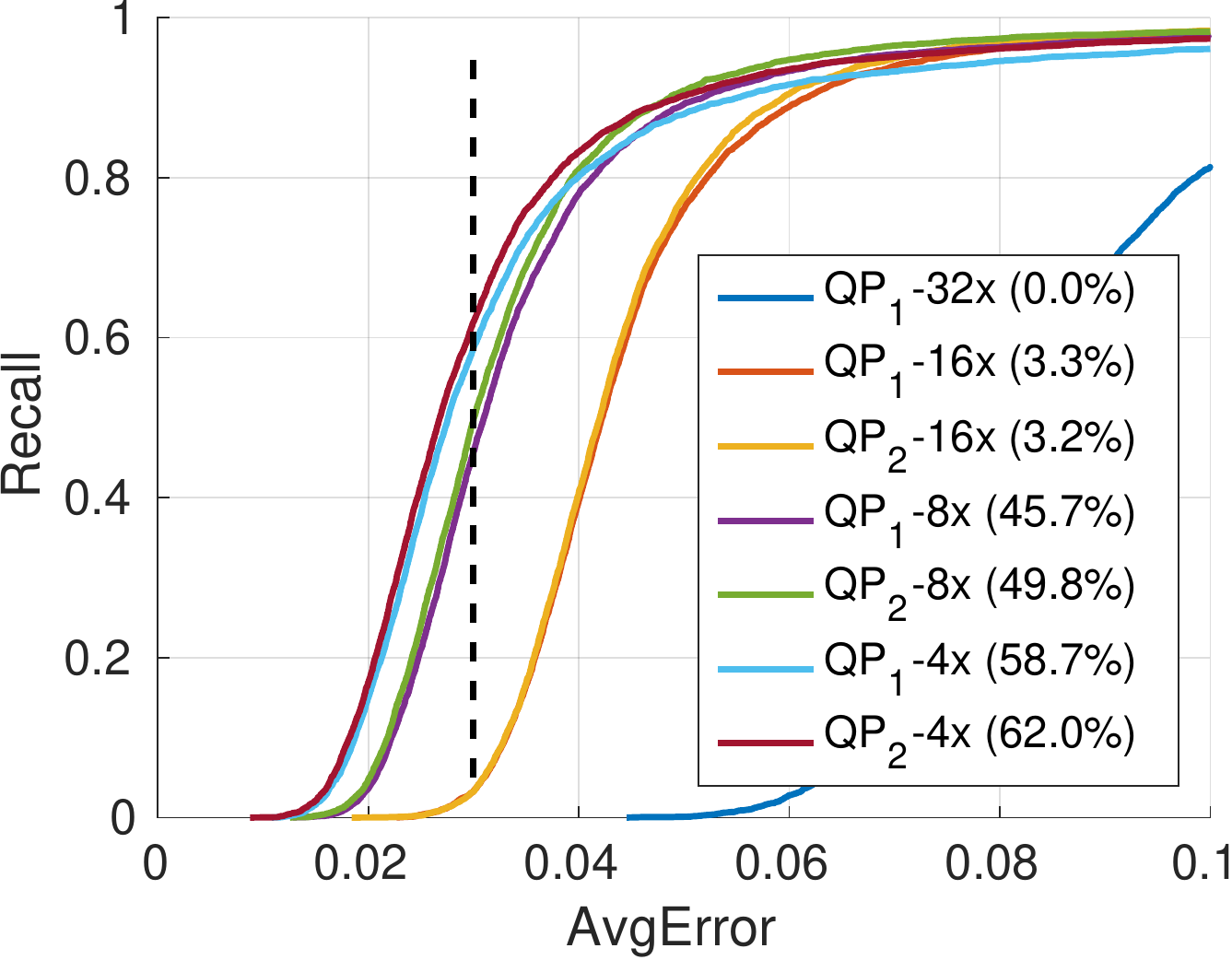}
  \caption{We plot the fraction of recalled face images whose average
    pixel localization error in AFLW (normalized by face
    size~\cite{zhu2012face}) is below a threshold (x-axis). We compare
    our \QP{1} and \QP{2} with varying numbers of scales used for
    multi-scale prediction, following the naming convention of
    FCN~\cite{long2014fully} (where the $Nx$ encodes the upsampling
    factor needed to resize the predicted heatmap to the original
    image resolution.) Single-scale models (\QP{1}-32x and \QP{2}-32x)
    are identical but perform quite poorly, not localizing any
    keypoints with 3.0\% of the face size. Adding more scales
    dramatically improves performance, and moreover, as we add
    additional scales, the relative improvement of \QP{2} also
    increases (as finer-scale features benefit the most from
    feedback). We visualize such models in Fig.~\ref{fig:coarse2fine}.}
  \label{fig:aflw-curve}
\end{figure}

{\bf AFLW:} The AFLW dataset \cite{kostinger2011annotated} is a
large-scale real-world collection of 25,993 faces in 21,997 real-world
images, annotated with facial keypoints. Notably, these faces are not
limited to be responses from an existing face detector, and so this
dataset contains more pose variation than other landmark datasets. We
hypothesized that such pose variation might illustrate the benefit of
bidirectional reasoning. Due to a lack of standard splits, we randomly
split the dataset into training (60\%), validation (20\%) and test
(20\%). 
As this is not a standard benchmark dataset, we compare to ourselves
for exploring the best practices to build multi-scale predictors for
keypoint localization (Fig.~\ref{fig:aflw-curve}). We include qualitative visualizations in Fig.~\ref{fig:aflw}.

\begin{figure}[t!]
  \centering
  \includegraphics[width=\linewidth]{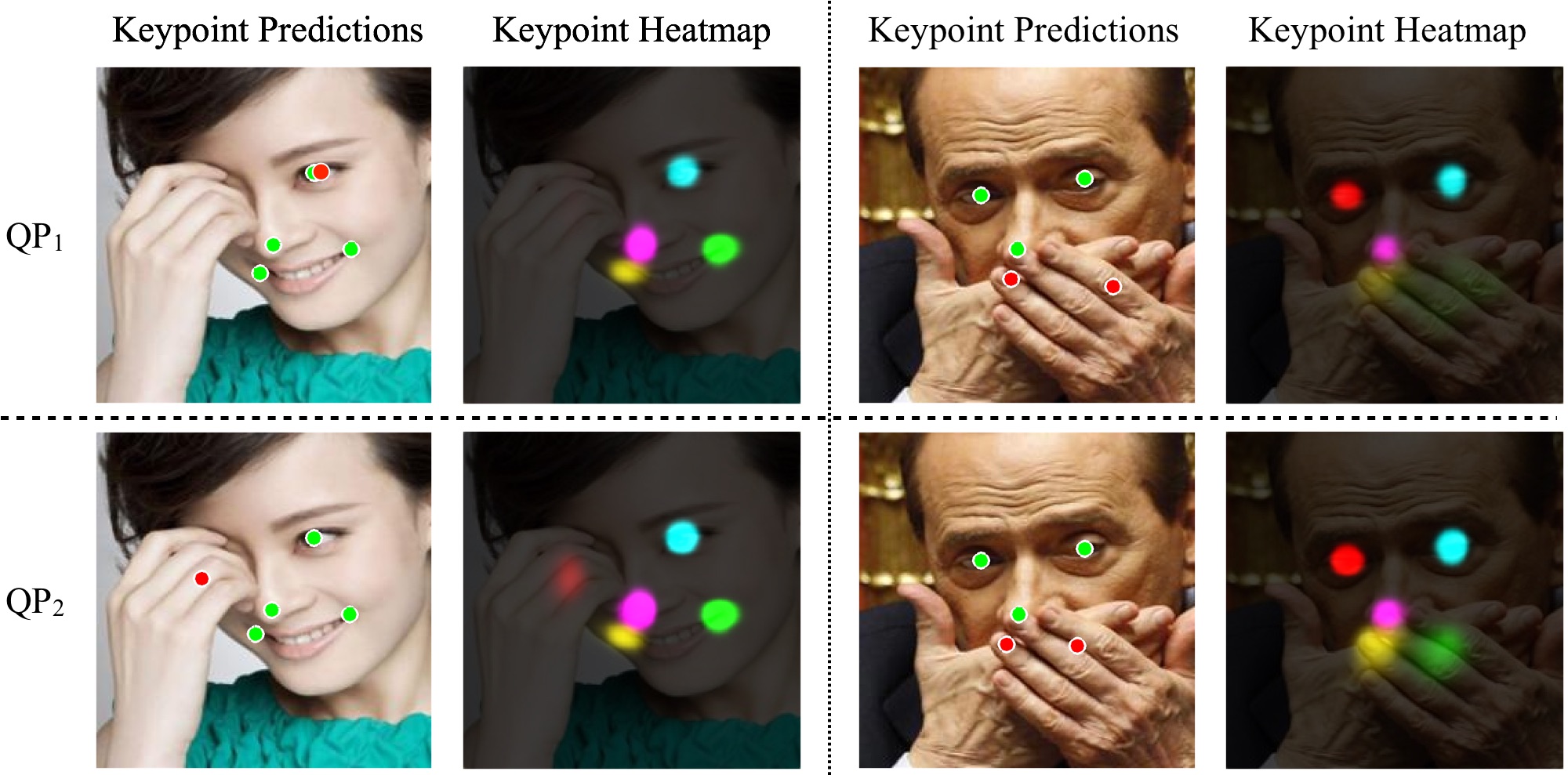}
  \caption{Visualization of keypoint predictions by \QP{1} and
    \QP{2} on two example COFW images. Both our models predict both
    keypoint locations and their visibility (produced by thresholding
    the value of the heatmap confidence at the predicted location). We
    denote (in)visible keypoint predictions with (red)green dots, and
    also plot the raw heatmap prediction as a colored distribution
    overlayed on a darkened image. Both our models correctly estimate
    keypoint visibility, but our bottom-up models \QP{1} misestimate
    their locations (because bottom-up evidence is misleading during
    occlusions). By integrating top-down knowledge (perhaps encoding
    spatial constraints on configurations of keypoints), \QP{2} is
    able to correctly estimate their locations.}
  \label{fig:cofw-heatmap}
\end{figure}

\begin{figure}[t!]
  \centering
  \begin{subfigure}{.75\linewidth}
    \includegraphics[width=\linewidth]{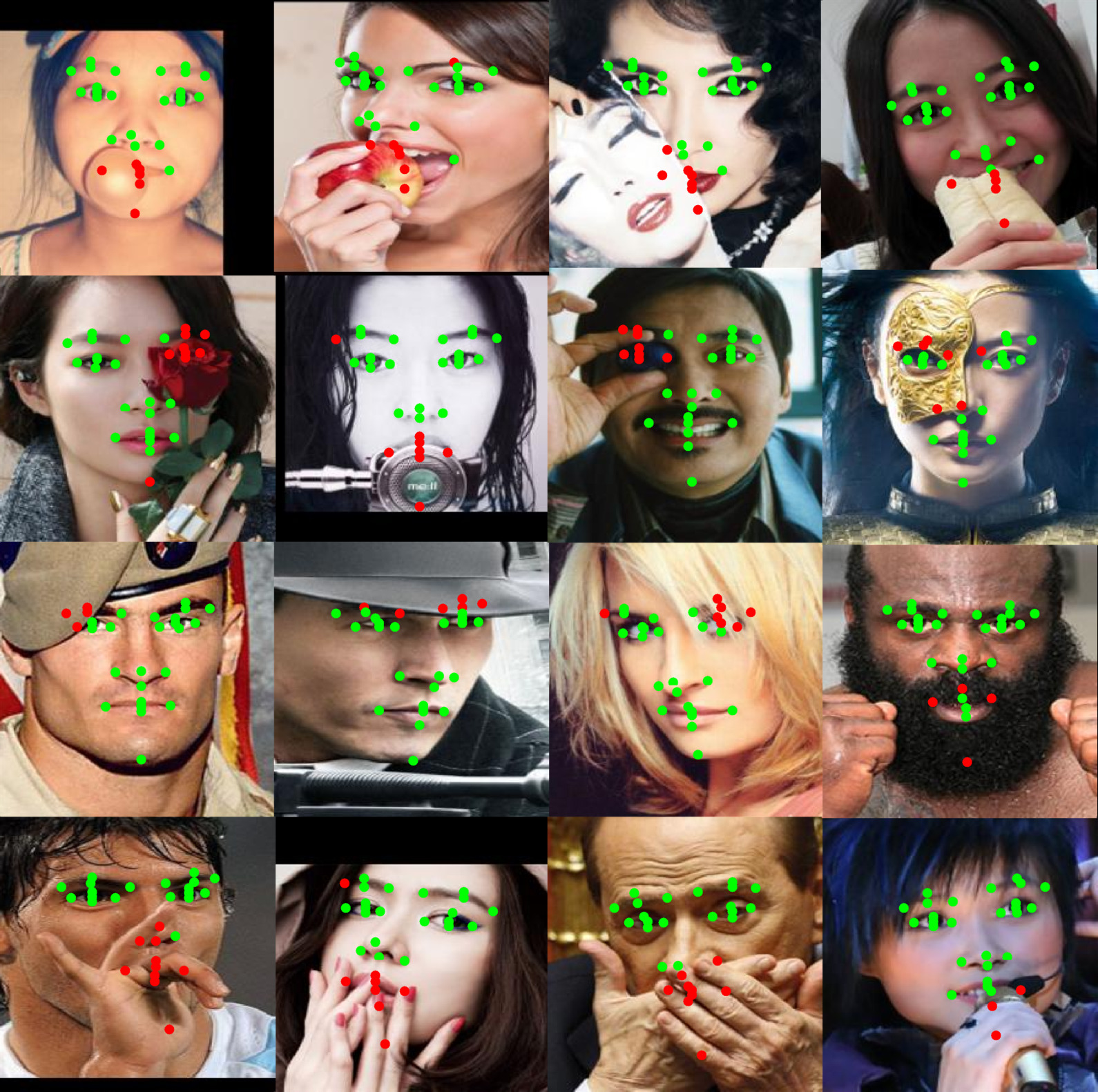}
    \caption{}
  \end{subfigure}
  \begin{subfigure}{.1925\linewidth}
    \includegraphics[width=\linewidth]{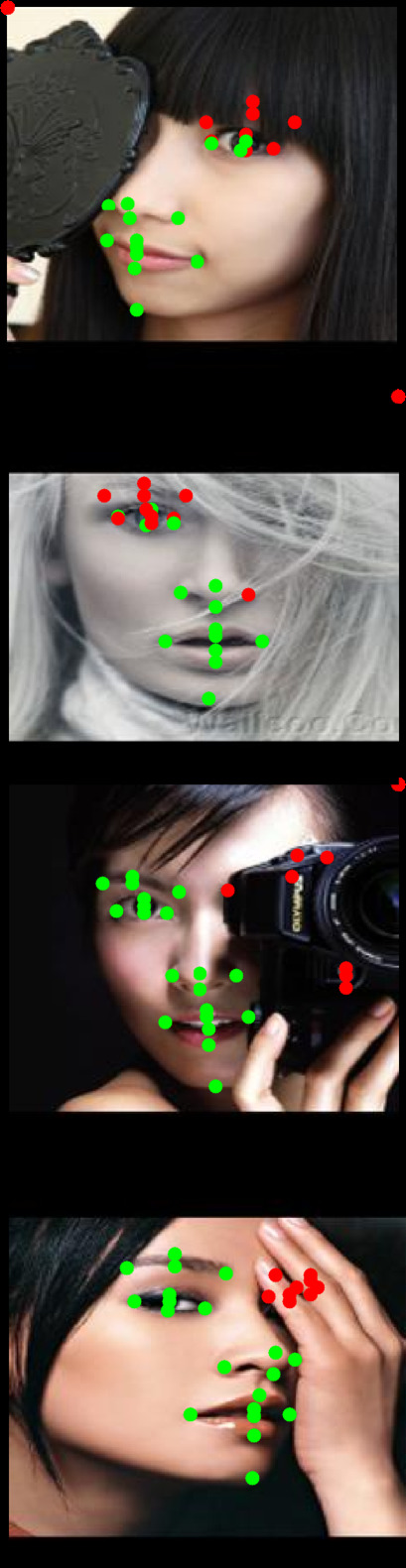}
    \caption{}
  \end{subfigure}
  \caption{Facial landmark localization and occlusion prediction
    results of \QP{2} on COFW, where red means occluded. Our bidirectional
    model is robust to occlusions caused by objects, hair, and
    skin. We also show cases where the model correctly predicts visibility but fails to accurately localize occluded landmarks ({\bf b}).}
  \label{fig:cofw}
\end{figure}

\begin{figure}[t]
  \centering
\includegraphics[width=.8\linewidth]{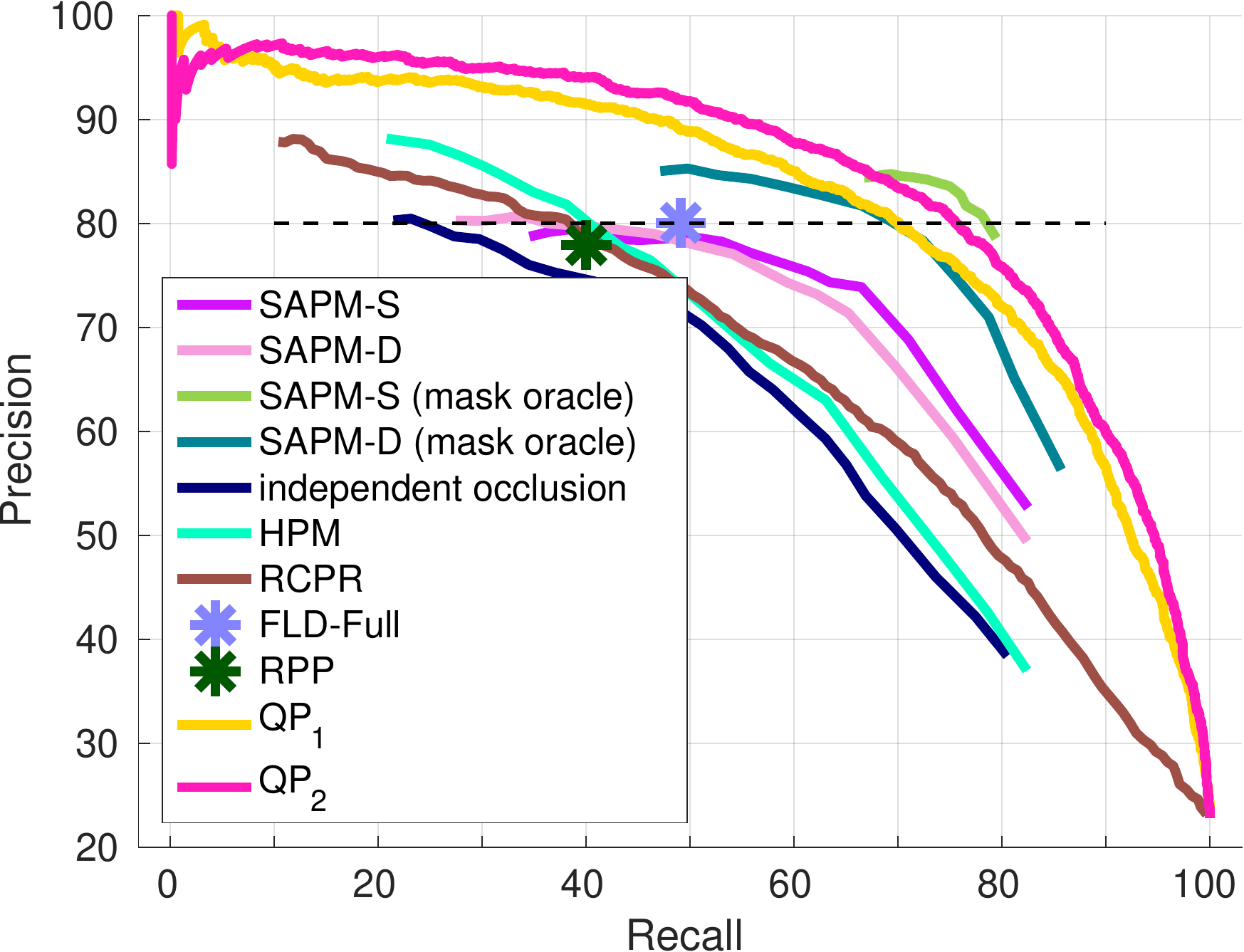}
  \caption{Keypoint visibility prediction on COFW, measured by
    precision-recall. Our bottom-up model \QP{1} already
    outperforms all past work that does not make use of ground-truth
    segmentation masks (where acronyms correspond those in
    Table~\ref{table:cofw}). Our top-down model \QP{2} even
    approaches the accuracy of such upper bounds. Following standard protocol,
    we evaluate and visualize accuracy in Fig.~\ref{fig:cofw} at a
    precision of 80\%. At such a level, our recall (76\%) significantly outperform the best
    previously-published recall of FLD~\cite{wu2015fld} (49\%).}
  \label{fig:cofw-curves}
\end{figure}

{\bf COFW:} Caltech Occluded Faces-in-the-Wild
(COFW)~\cite{burgos2013robust} is dataset of 1007 face images with
severe occlusions. We present qualitative results in
Fig.~\ref{fig:cofw-heatmap} and Fig.~\ref{fig:cofw}, and quantitative
results in Table~\ref{table:cofw} and Fig.~\ref{fig:cofw-curves}. Our
bottom-up \QP{1} already performs near the state-of-the-art, while the
\QP{2} significantly improves in accuracy of visible landmark
localization and occlusion prediction. In terms of the latter, our
model even approaches upper bounds that make use of ground-truth
segmentation labels~\cite{ghiasi2015sapm}. Our models are not quite
state-of-the-art in localizing occluded points. We believe this may
point to a limitation in the underlying benchmark.  Consider an image
of a face mostly occluded by the hand
(Fig.~\ref{fig:cofw-heatmap}). In such cases, humans may not even
agree on keypoint locations, indicating that a keypoint {\em
  distribution} may be a more reasonable target output. Our models
provide such uncertainty estimates, while most keypoint architectures
based on regression cannot.

\begin{table}
  \begin{center}
    \begin{tabular}{l|c|c}
                                    & Visible Points & All Points \\
      \hline \hline
      RCPR\cite{burgos2013robust}   & -              & 8.5        \\
      RPP\cite{yang2015robust}      & -              & 7.52       \\
      HPM\cite{ghiasi2014occlusion} & -              & 7.46       \\
      SAPM\cite{ghiasi2015sapm}     & 5.77           & 6.89       \\
      FLD-Full\cite{wu2015fld}      & 5.18           & {\bf 5.93} \\
      \hline
      \QP{1}                        & 5.26           & 10.06      \\
      \QP{2}                        & {\bf 4.67}     & 7.87       \\
    \end{tabular}
  \end{center}
  \caption{Average keypoint localization error (as a fraction of
    inter-ocular distance) on COFW. When adding top-down feedback
    (\QP{2}), our accuracy on visible keypoints significantly improves
    upon prior work. In the text, we argue that such localization
    results are more meaningful than those for occluded keypoints. In
    Fig.~\ref{fig:cofw-curves}, we show that our models significantly
    outperform all prior work in terms of keypoint visibility
    prediction. }
  \label{table:cofw}
\end{table}

\begin{figure}[t!]
  \centering
  \includegraphics[width=.8\linewidth]{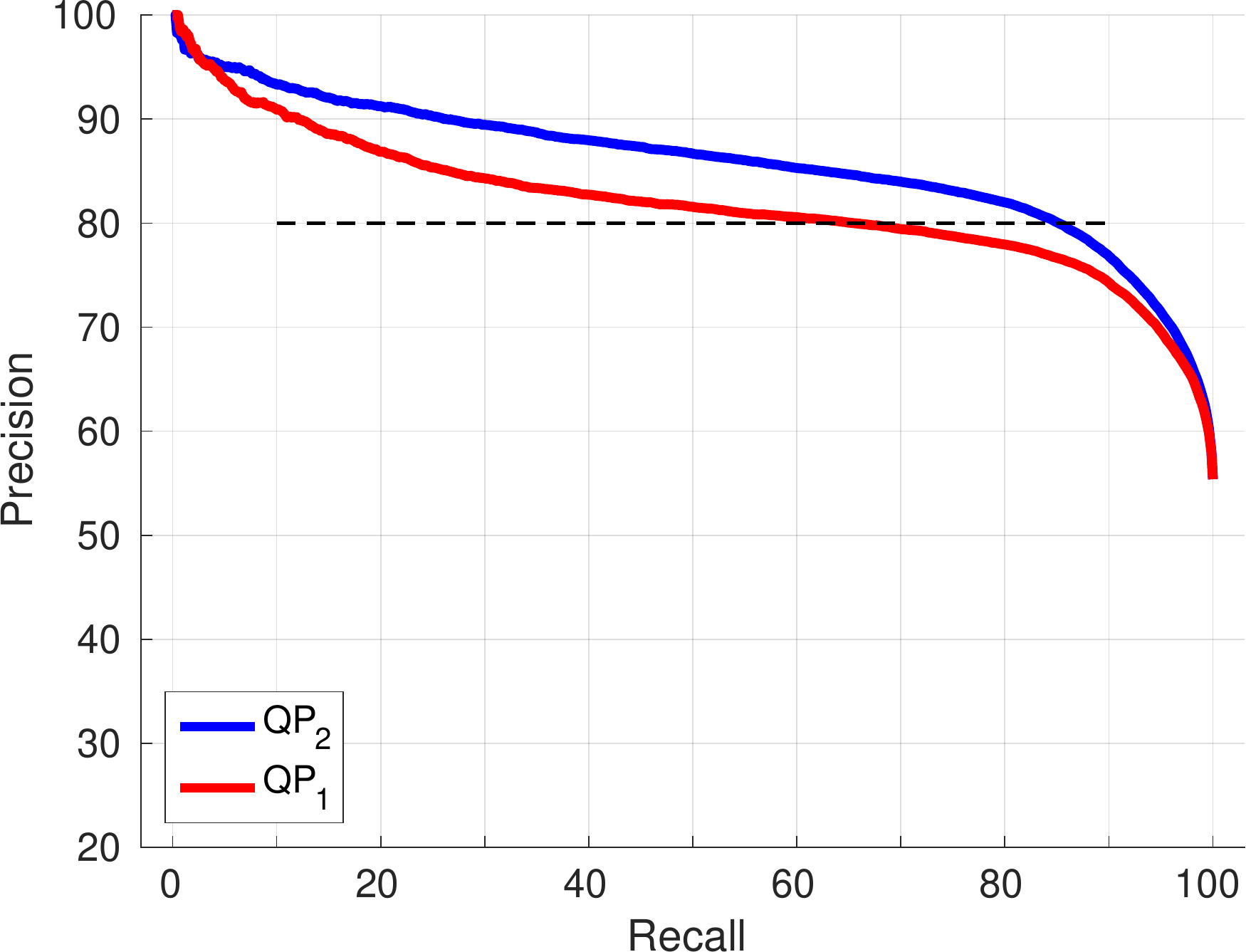}
  \caption{Keypoint visibility prediction on Pascal Person (a dataset with significant occlusion and truncation), measured by
    precision-recall curves. At 80\% precision, our top-down model ($QP_2$) significantly improves recall from 65\% to 85\%.}
  \label{fig:pascal-occ}
\end{figure}

{\bf Pascal Person:} The Pascal 2011 Person
dataset~\cite{hariharan2011semantic} consists of 11,599 person
instances, each annotated with a bounding box around the visible
region and up to 23 human keypoints per person. This dataset contains
significant occlusions. 
We follow the evaluation protocol of~\cite{long2014convnets} and
present results for localization of visible keypoints on a standard
testset in Table~\ref{table:pascal}. Our bottom-up \QP{1} model
already significantly improves upon the state-of-the-art (including
prior work making use of deep features), while our top-down models
\QP{2} further improve accuracy by 2\% without any increase in model
complexity (as measured by the number of parameters). Note that the
standard evaluation protocols evaluate only visible keypoints. In
Fig.~\ref{fig:pascal-occ}, we demonstrate that our model can also
accurately predict keypoint visibility ``for free''.

\begin{figure*}[t!]
  \centering
  \includegraphics[width=.9\linewidth]{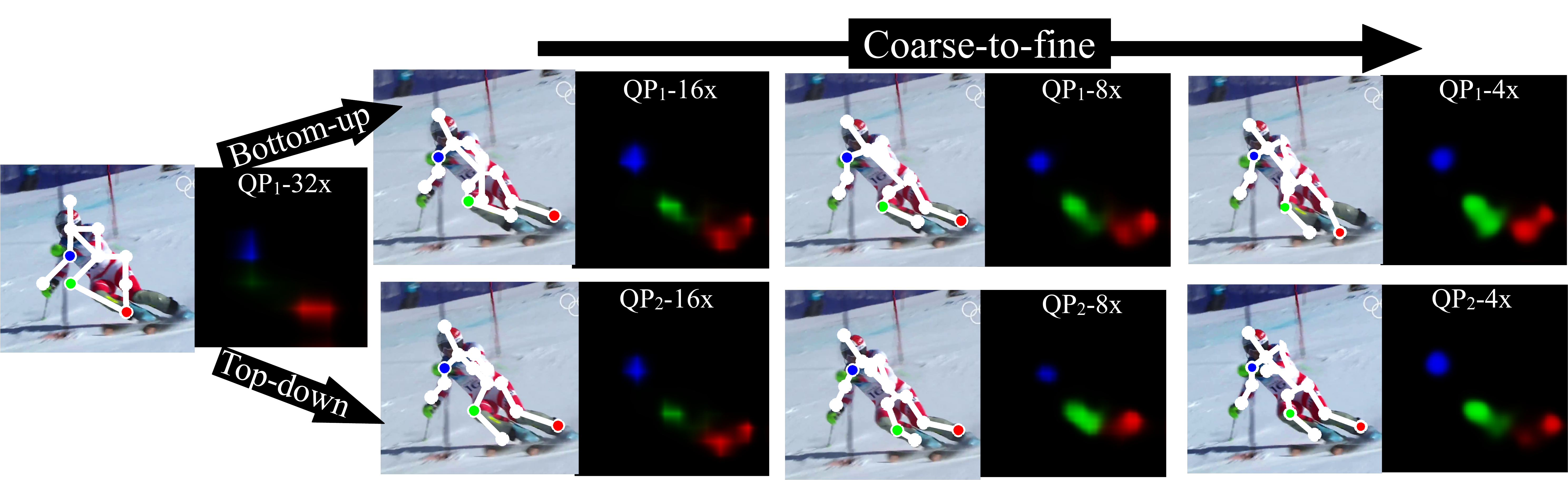}
  \caption{We visualize bottom-up and top-down models trained for human pose estimation, using the naming convention of Fig.~\ref{fig:aflw-curve}. Top-down feedback (\QP{2}) more accurately guides finer-scale predictions, resolving left-right ambiguities in the ankle (red) and poor localization of the knee (green) in the bottom-up model (\QP{1}).}
  \label{fig:coarse2fine}
\end{figure*}

\begin{table}
  \centering
    \begin{tabular}{l|c|c}
      $\alpha$                         & 0.10       & 0.20       \\
      \hline \hline
      CNN+prior\cite{long2014convnets} & 47.1       & -          \\
      \hline
      \QP{1}                           & 66.5       & 78.9       \\
      \QP{2}                           & {\bf 68.8} & {\bf 80.8} \\
    \end{tabular}
    \caption{We show human keypoint localization performance on PASCAL
      VOC 2011 Person following the evaluation protocol in
      \cite{long2014convnets}. PCK refers to the fraction of keypoints
      that were localized within some distance (measured with respect
      to the instance's bounding box). Our bottom-up models already
      significantly improve results across all distance thresholds
      ($\alpha = 10,20\%$). Our top-down models add a 2\% improvement
      without increasing the number of parameters. }
    \label{table:pascal}
\end{table}

\begin{figure}[t!]
  \centering
  \includegraphics[width=.8\linewidth]{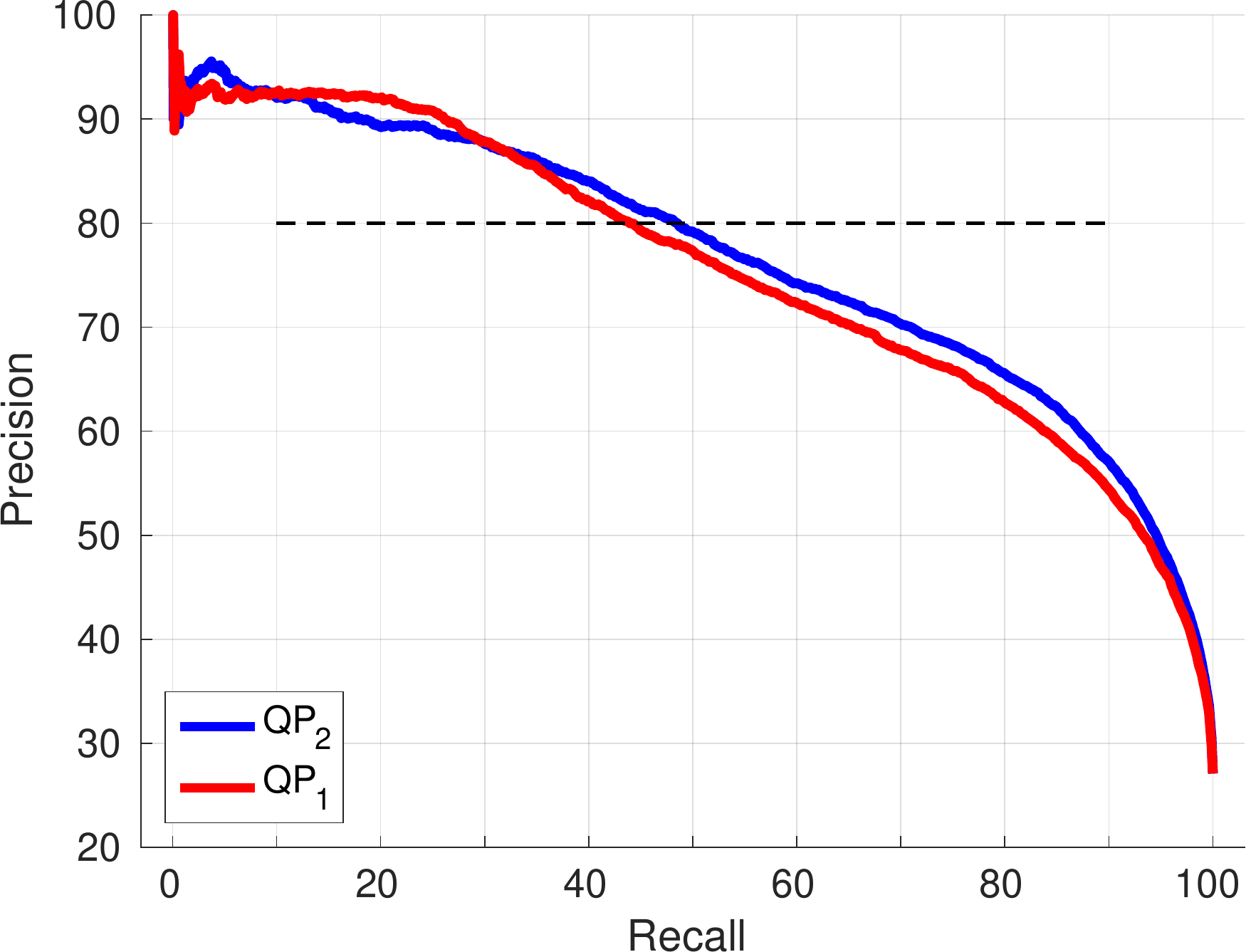}
  \caption{Keypoint visibility prediction on MPII, measured by
    precision-recall curves. At 80\% precision, our top-down model ($QP_2$) improves recall from 44\% to 49\%.}
  \label{fig:mpii-occ}
\end{figure}

\begin{figure}[t]
  \centering
  \includegraphics[width=.95\linewidth]{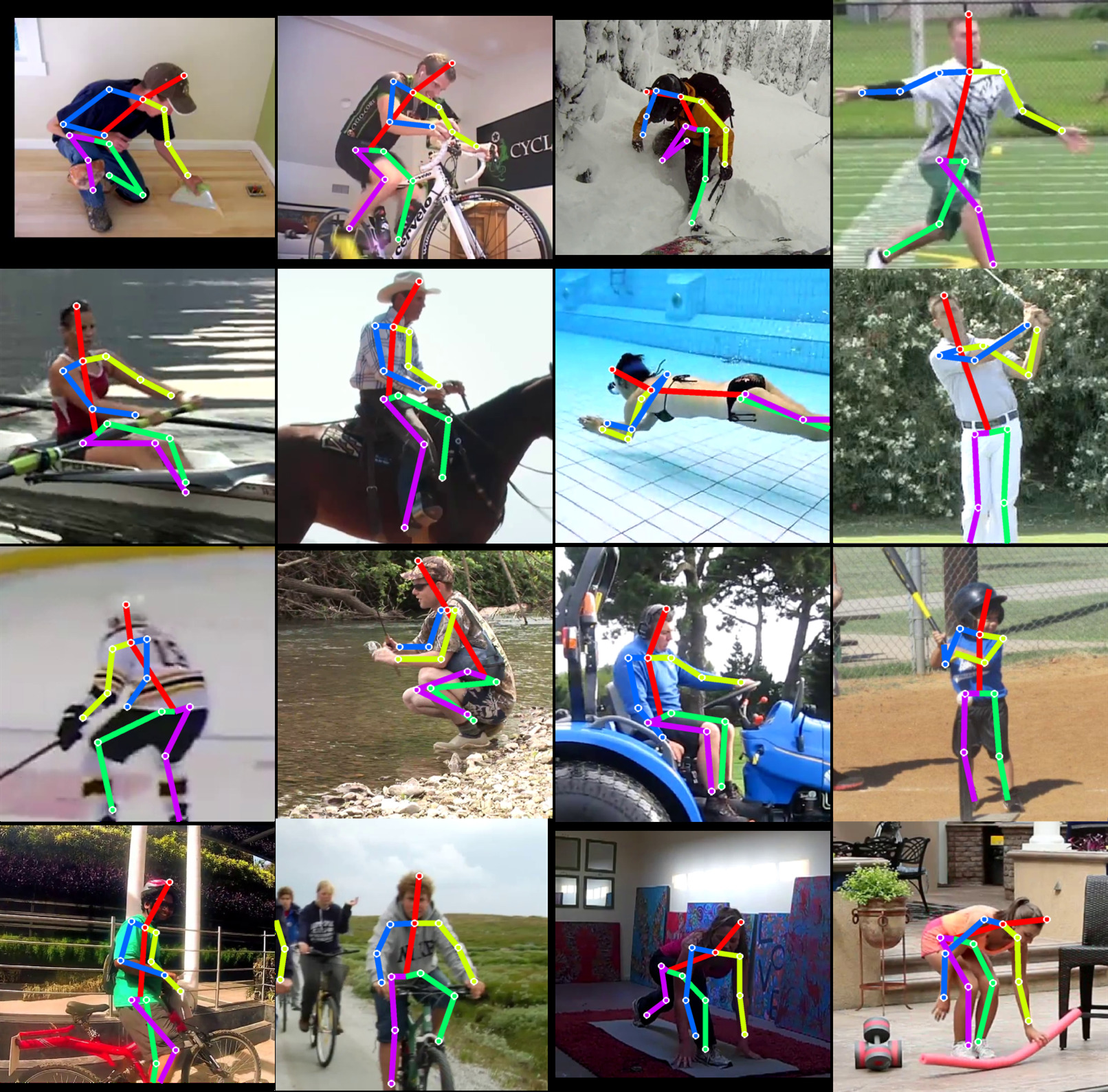}
\caption{Keypoint localization results of \QP{2} on the MPII Human
  Pose testset. We quantitatively evaluate results on the validation
  set in Table~\ref{table:pascal}. Our models are able to localize
  keypoints even under significant occlusions. Recall that our models
  can also predict visibility labels ``for free'', as shown in
  Fig.~\ref{fig:mpii-occ}.}
  \label{fig:mpii}
\end{figure}

{\bf MPII:} MPII is (to our knowledge) the largest available
articulated human pose dataset~\cite{andriluka14cvpr}, consisting of
40,000 people instances annotated with keypoints, visibility flags,
and activity labels. We present qualitative results in
Fig.~\ref{fig:mpii} and quantitative results in
Table~\ref{table:mpii}. Our top-down model \QP{2} appears to
outperform all prior work on full-body keypoints. Note that this dataset also includes visibility labels for keypoints, even though these are not
part of the standard evaluation protocol. In Fig.~\ref{fig:mpii-occ},
we demonstrate that visibility prediction on MPII also benefits from
top-down feedback.

{\bf TB: } It is worth contrasting our results with TB~\cite{tompson2015efficient}, which implicitly models feedback by (1) using a MRF to post-process CNN outputs to ensure
kinematic consistency between keypoints and (2) using high-level
predictions from a coarse CNN to adaptively crop high-res features for a fine CNN. Our single CNN endowed with
top-down feedback is slightly more accurate without requiring any
additional parameters, while being 2X faster (86.5 ms vs TB's 157.2
ms). These results suggest that top-down reasoning may elegantly
capture structured outputs and attention, two active areas of research
in deep learning.

\renewcommand{\tabcolsep}{2pt}
\begin{table}[t!]
  \centering
  \resizebox{\linewidth}{!}{
    \begin{tabular}{l|c|c|c|c|c|c|c||c|c}
                                        & Head & Shou & Elb  & Wri  & Hip  & Kne  & Ank  & Upp  & Full \\
      \hline
      \hline
      GM \cite{gkioxari2013articulated} & -    & 36.3 & 26.1 & 15.3 & -    & -    & -    & 25.9 & -    \\
      ST \cite{sapp2013modec}           & -    & 38.0 & 26.3 & 19.3 & -    & -    & -    & 27.9 & -    \\
      YR \cite{yang2011articulated}     & 73.2 & 56.2 & 41.3 & 32.1 & 36.2 & 33.2 & 34.5 & 43.2 & 44.5 \\
      PS \cite{pishchulin2013poselet}   & 74.2 & 49.0 & 40.8 & 34.1 & 36.5 & 34.4 & 35.1 & 41.3 & 44.0 \\
      TB \cite{tompson2015efficient}    & 96.1 & 91.9 & 83.9 & 77.8 & 80.9 & 72.3 & 64.8 & {\bf 84.5} & 82.0 \\
      \hline
      \QP{1}                            & 94.3 & 90.4 & 81.6 & 75.2 & 80.1 & 73.0 & 68.3 & 82.4 & 81.1 \\
      \QP{2}                            & 95.0 & 91.6 & 83.0 & 76.6 & 81.9 & 74.5 & 69.5 & 83.8 & {\bf 82.4} \\
    \end{tabular}
  }
  \caption{We show PCKh-0.5 keypoint localization results on MPII
    using the recommended benchmark protocol~\cite{andriluka14cvpr}.} 
  \label{table:mpii}
\end{table}

{\bf More recurrence iterations: } To explore \QP{K}'s performance as
a function of $K$ without exceeding memory limits, we trained a smaller
network from scratch on 56X56 sized inputs for 100 epochs. As shown in
Table~\ref{fig:mpii-small}, we conclude: (1) all recurrent models
outperform the bottom-up baseline \QP{1}; (2) additional iterations
generally helps, but performance maxes out at \QP{4}. A two-pass model
(\QP{2}) is surprisingly effective at capturing top-down info while
being fast and easy to train. 

\renewcommand{\tabcolsep}{2pt}
\begin{table}[t!]
  \centering
    \begin{tabular}{|c|c|c|c|c|c|c|}
      \hline 
      K          & 1    & 2    & 3    & 4             & 5    & 6    \\ 
      \hline 
      Upper Body & 57.8 & 59.6 & 58.7 & {\bf 61.4} & 58.7 & 60.9 \\ 
      \hline 
      Full Body  & 59.8 & 62.3 & 61.0 & {\bf 63.1} & 61.2 & 62.6 \\ 
      \hline 
    \end{tabular}
  \caption{ PCKh(.5) on MPII-Val for a smaller network } 
  \label{fig:mpii-small}
\end{table}


{\bf Conclusion:} We show that hierarchical Rectified Gaussian models
can be optimized with rectified neural networks. From a modeling
perspective, this observation allows one to discriminatively-train
such probabilistic models with neural toolboxes. From a neural net
perspective, this observation provides a theoretically-elegant
approach for endowing CNNs with top-down feedback -- {\em without any
  increase in the number of parameters}. To thoroughly evaluate our
models, we focus on ``vision-with-scrutiny'' tasks such as keypoint
localization, making use of well-known benchmark datasets. We
introduce (near) state-of-the-art bottom-up baselines based on
multi-scale prediction, and consistently improve upon those results
with top-down feedback (particularly during occlusions when bottom-up
evidence may be ambiguous).

{\bf Acknowledgments:} This research is supported by NSF
Grant 0954083 and by the Office of the Director
of National Intelligence (ODNI), Intelligence Advanced Research
Projects Activity (IARPA), via IARPA R \& D Contract
No. 2014-14071600012. The views and conclusions contained herein are
those of the authors and should not be interpreted as necessarily
representing the official policies or endorsements, either expressed
or implied, of the ODNI, IARPA, or the U.S. Government. The
U.S. Government is authorized to reproduce and distribute reprints for
Governmental purposes notwithstanding any copyright annotation
thereon.


{\small
  \bibliographystyle{ieee}
  \bibliography{ref}
}

\end{document}